\documentclass{article}

\usepackage{times}
\usepackage{setspace}

\usepackage[dvips]{graphicx} % when using pdfLatex (preferred)

\begin{document}
\begin{titlepage}
\title{The Cyborg Astrobiologist: Scouting Red Beds for Uncommon Features with Geological Significance}
\author{
\and
Patrick Charles McGuire\footnote{corresponding author: {\it Email:} mcguire@physik.uni-bielefeld.de\hspace{12cm}\break{\it Telephone:} +34 91 520 6432 {\it Fax:} +34 91 520 1621\hspace{12cm}\break
\vspace{0.1cm}\break
Publication information: Received 4 April 2005, accepted in revised form 20 May 2005,\hspace{10cm}\break
to appear in the {\it International Journal of Astrobiology}, vol. {\bf 4}, issue 2 (June 2005)} \\
{\it Robotics \& Planetary Exploration Laboratory, and Transdisciplinary Laboratory}\\
{\it Centro de Astrobiolog\'ia (INTA/CSIC)}\\
{\it Instituto Nacional T\'ecnica Aeroespacial}\\
{\it Carretera de Torrej\'on a Ajalvir km 4.5, Torrej\'on de Ardoz, Madrid, Spain 28850}\\
\and
Enrique D\'iaz-Mart\'inez\\
{\it (currently at) Direcci\'on de Geolog\'ia y Geof\'isica}\\
{\it Instituto Geol\'ogico y Minero de Espa\~na}\\
{\it Calera 1, Tres Cantos, Madrid, Spain 28760} \\
{\it (formerly at) Planetary Geology Laboratory, Centro de Astrobiolog\'ia}\\
\and
Jens Orm\"o\\
{\it Planetary Geology Laboratory}\\
{\it Centro de Astrobiolog\'ia (INTA/CSIC), Instituto Nacional T\'ecnica Aeroespacial}\\
{\it Carretera de Torrej\'on a Ajalvir km 4.5, Torrej\'on de Ardoz, Madrid, Spain 28850}\\
\and
Javier G\'omez-Elvira, Jos\'e Antonio Rodriguez-Manfredi, Eduardo Sebasti\'an-Mart\'inez\\
{\it Robotics \& Planetary Exploration Laboratory}\\
{\it Centro de Astrobiolog\'ia (INTA/CSIC), Instituto Nacional T\'ecnica Aeroespacial}\\
{\it Carretera de Torrej\'on a Ajalvir km 4.5, Torrej\'on de Ardoz, Madrid, Spain 28850}\\
\and
Helge Ritter, Robert Haschke, Markus Oesker, J\"org Ontrup\\
{\it Neuroinformatics Group, Computer Science Department}\\
{\it Technische Fakult\"at}\\
{\it University of Bielefeld}\\
{\it P.O.-Box 10 01 31}\\
{\it Bielefeld, Germany 33501}\\
\date{\today}
\vspace{3.0cm}
}
\maketitle
\thispagestyle{empty}
\end{titlepage}
\pagenumbering{arabic}
\begin{abstract}
The `Cyborg Astrobiologist' has undergone a second geological field trial, at a site in northern Guadalajara, Spain, near Riba de Santiuste. The site at Riba de Santiuste is dominated by layered deposits of red sandstones. The Cyborg Astrobiologist is a wearable computer and video camera system that has demonstrated a capability to find uncommon interest points in geological imagery in real-time in the field. In this second field trial, the computer vision system of the Cyborg Astrobiologist was tested at seven different tripod positions, on three different geological structures. The first geological structure was an outcrop of nearly homogeneous sandstone, which exhibits oxidized-iron impurities in red and and an absence of these iron impurities in white. The white areas in these ``red beds'' have turned white because the iron has been removed. The iron removal from the sandstone can proceed once the iron has been chemically reduced, perhaps by a biological agent.  The computer vision system found in one instance several (iron-free) white spots to be uncommon and therefore interesting, as well as several small and dark nodules. The second geological structure was another outcrop some 600 meters to the east, with white, textured mineral deposits on the surface of the sandstone, at the bottom of the outcrop. The computer vision system found these white, textured mineral deposits to be interesting. We acquired samples of the mineral deposits for geochemical analysis in the laboratory. This laboratory analysis of the crust identifies a double layer, consisting of an internal millimeter-size layering of calcite and an external centimeter-size effluorescence of gypsum. The third geological structure was a 50 cm thick paleosol layer, with fossilized root structures of some plants. The computer vision system also found certain areas of these root structures to be interesting. A quasi-blind comparison of the Cyborg Astrobiologist's interest points for these images with the interest points determined afterwards by a human geologist shows that the Cyborg Astrobiologist concurred with the human geologist 68\% of the time (true positive rate), with a 32\% false positive rate and a 32\% false negative rate. The performance of the Cyborg Astrobiologist's computer vision system was by no means perfect, so there is plenty of room for improvement. However, these tests validate the image-segmentation and uncommon-mapping technique that we first employed at a different geological site (Rivas Vaciamadrid) with somewhat different properties of the imagery.
\end{abstract}

{\bf Keywords:} computer vision, robotics, image segmentation, uncommon map, interest map, field geology, Mars, Meridiani Planum, wearable computers, co-occurrence histograms, red beds, red sandstones, nodules, concretions, reduction spheroids, Triassic period.
\markright{The Cyborg Astrobiologist: Scouting Red Beds in Guadalajara: McGuire et al.}\
\thispagestyle{myheadings}
\newpage

\markright{The Cyborg Astrobiologist: Scouting Red Beds in Guadalajara: McGuire et al.}\

\begin{spacing}{1}
\section{Introduction}

\markright{The Cyborg Astrobiologist: Scouting Red Beds in Guadalajara}

\begin{figure}
 \center{\includegraphics[height=8.0cm]{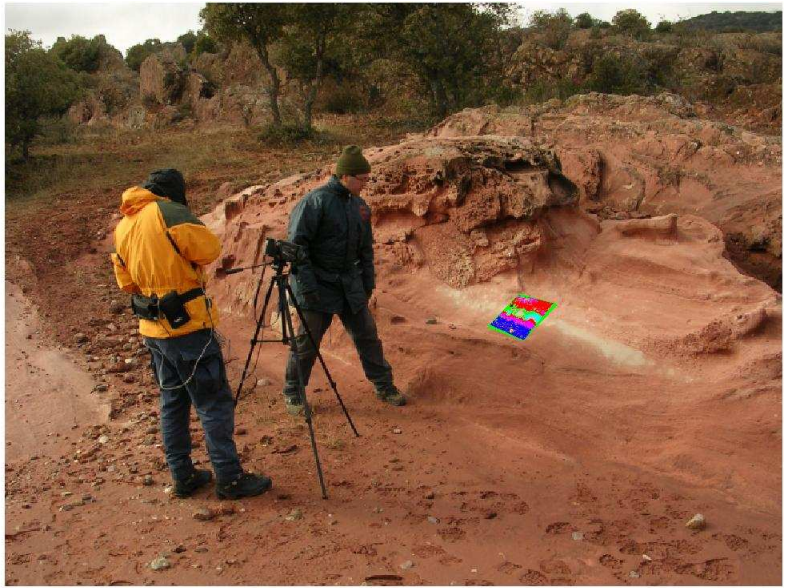}}
  \caption{ Orm\"o with the Cyborg Astrobiologist system at Tripod Position \#2, near Outcrop \#1 outside of Riba de Santiuste in the northern part of the Guadalajara province in central Spain. D\'iaz Mart\'inez is assisting; he is trying to identify the positions of the computer-determined interest points on the outcrop, which are annotated on the tablet display of the wearable computer. The image was segmented in real time, and we overlay the image segmentation (for saturation) on the red sandstone outcrop. In Fig.\ref{Explanation}, we present details of the determination of the interest map and the interest points for the image acquired and processed at this tripod position.
This documents the Riba mission at 1:53PM on 8 February 2004.}
\label{EnriqueJens}
 \end{figure}

Half a lifetime has passed since human astronauts directly explored another planetary body. Currently, two robotic
geologists are exploring the surface of Mars, one scientific probe has just landed on Saturn's moon Titan, and a number of orbiters are studying several of the planets and moons in our solar system. Soon, we will be sending
more robotic explorers to Mars, more science stations to orbit our planetary neighbors, and perhaps new human explorers to the Moon. All of these exploration systems, human or robotic, semi-autonomous or remote-operated, can benefit from enhancements in their capabilities of scientific autonomy.
Human astronaut explorers can benefit from enhanced scientific autonomy -- astronauts with ``augmented reality'' visors could explore more efficiently and perhaps make more discoveries than astronauts who exclusively rely upon guidance from Earth-based scientists and engineers. Remote-operated robotic rovers can benefit from enhanced scientific autonomy -- this autonomy could either be on-board the rover\footnote{i.e., in the form of `macros' or callable subroutines.} or in computers on the Earth\footnote{i.e. in order to assist the night-time mission planners.}. Such scientific autonomy can enhance the scientific productivity of these expensive missions, but only if the autonomy measures are well tested, and therefore `trustable' by the controllers on the Earth.

Along these lines, we have constructed a field-capable platform (at relatively low cost) in order to develop and to test computer-vision algorithms for space exploration here on the Earth, prior to deployment in a space mission. This platform uses a wearable computer and a video camera\footnote{We are currently working towards extending these capabilities with the addition of a microscopic imager. In the future, we could also integrate near-infrared or mid-infrared spectrometers into the system, and perhaps a Raman spectrometer. For example, a thermal-emission spectrometer (TES) would be useful to identify minerals in the field.}, as well as the human operator. Since this astrobiological exploration system is part human and part machine, we call our system the `Cyborg Astrobiologist'\footnote{The word `Cyborg' is derived from `{\it Cyb}ernetic {\it Org}anism'.}. Recently, we have reported our first results at a geological field site with the Cyborg Astrobiologist exploration system (McGuire {\it et al.}\ 2004b). In this paper, we discuss our results at a second geological field site, near Riba de Santiuste, in northern Guadalajara, Spain (see the maps in Figures 2 \& 3). This second geological field site offers a different type of imagery than was studied in the previous paper. This new imagery resembles some aspects of the imagery that Mars Exploration Rover (MER) Opportunity is currently studying on Mars, and the new imagery has greater astrobiological implications than the imagery studied in the previous mission of the Cyborg Astrobiologist. We show that the basic exploration algorithms that we introduced in our first paper also function rather well at this second field site, despite the change in nature of the imagery.

To give context to our work, we would like to remind the reader of two discoveries by the Apollo 15 and the Apollo 17 astronauts on the Moon in the early 1970's, namely of the `Genesis Rock' and of `orange soil' (Compton 1989). The Apollo 15 astronauts were ``astronauts trained to be geologists", and one of their missions was to find `ancient' rocks, in order to obtain information as to how the Moon was formed. Given the `bias' of their mission, and given significant `scientific autonomy' by mission control on the second day of their mission at the Hadley-Appenines landing site, astronauts Scott \& Irwin found an anorthosite\footnote{Anorthosite has a calcium aluminum silicate composition, with a crystalline and `plagioclase' form.  A plagioclase is a rock that exhibits obliquely-cleaving crystals}. Scott \& Irwin dubbed this specimen `the Genesis rock', because it possibly recorded the time after the molten Moon's surface first cooled down, over 4 billion years ago. They initially found the Genesis rock to be interesting partly because their mission requirements and their focused geological training biased them to look for such crystalline types of rocks.

 This biased-search during the Apollo 15 mission on the Moon is not unlike one mission requirement of the MER Opportunity rover. One main objective for sending Opportunity to the Meridiani Planum site on Mars was to understand the Mars Odyssey orbiter's suggestion of abundant coarse-grained gray hematite at Meridiani. So the initial focus and bias of Opportunity's study of Meridiani was on determining the source and nature of the hematite, and based upon this bias, Opportunity and the Earth-based geologists and engineers have been highly successful in writing `the story' about hematite at Meridiani (Squyres {\it et al.} 2004; Chan {\it et al.} 2004). 

 The Apollo 17 astronauts had one astronaut who had been trained to be a geologist (Cernan) and one geologist who had trained to be an astronaut (Schmitt). One of Schmitt's responsibilities (as geologist/astronaut) was to try to see if he could observe phenomena on the Moon which the previous astronaut/geologists had not seen. Another responsibility for both Cernan and Schmitt was to try to find signs of recent volcanism, which was one reason their Taurus-Littrow landing site had been chosen, since it contained numerous craters of possible volcanic origin. Despite the bias of the mission towards volcanic geology, Schmitt discovered an unusual `orange soil' on the rim of one crater, which he initially thought could be due to some unusual oxidative process. This orange soil later turned out to be composed of small orangish glass-like beads, probably of volcanic origin. As with the rest of Schmitt's geological observing on the Moon in which ``he based his decisions on taking samples on visually detectable differences or similarities" (Compton 1989), Schmitt (Schmitt 1987) found the orange soil interesting because it was different from anything else he had seen on the Moon. From our point of view, since Schmitt was an expert geologist, he was able to go much beyond the naive mission goals and biases of looking for fresh signs of volcanism, to discover something else (orange soil), which at the time seemed unrelated to volcanism. In summary, we point to the Apollo 15 \& 17 missions and the MER Opportunity mission as examples of the need to have both biased and unbiased techniques for scientific autonomy during space exploration  missions.

It is rather difficult to make a system that can reliably detect signatures of interesting geological and biological phenomena (with an imager or with a spectrometer) in a general and biased manner. In this report, we describe the further testing of our {\it unbiased} ``Cyborg Astrobiologist'' system.

In Section 2, we discuss the hardware and software of our system, followed by summaries \& results of the geological expeditions to Riba de Santiuste in Sections 3 \& 4, and finishing with more general discussion \& conclusions in Section 5.

\section{The Cyborg Geologist \& Astrobiologist System}
\markright{The Cyborg Astrobiologist: System}
 Our ongoing effort in the area of autonomous recognition of scientific targets-of-opportunity for field geology and field astrobiology is maturing. To date, we have developed and field-tested a ``Cyborg Astrobiologist'' system (McGuire {\it et al.}\ 2004a; McGuire {\it et al.}\ 2004b) that now can:
\begin{spacing}{1}
\begin{trivlist}
\raggedright
\item $\bullet$ Use human mobility to maneuver to and within a geological site;
\item $\bullet$ Use a portable robotic camera system to obtain a mosaic of color images;
\item $\bullet$ Use a `wearable' computer to search in real-time for the most uncommon regions of these mosaic images;
\item $\bullet$ Use the robotic camera system to repoint at several of the most uncommon areas of the mosaic images, in order to obtain much more detailed information about these `interesting' uncommon areas;
\item $\bullet$ Choose one of the interesting areas in the panorama for closer approach; and
\item $\bullet$ Repeat the process as often as desired, sometimes retracing a step of geological approach. 
\end{trivlist}
\end{spacing}

   In the Mars Exploration Workshop in Madrid in November 2003, we demonstrated some of the early capabilities of our `Cyborg' Geologist/Astrobiologist System (McGuire {\it et al.}\ 2004a). We have been using this Cyborg system as a platform to develop computer-vision algorithms for recognizing interesting geological and astrobiological features, and for testing these algorithms in the field here on the Earth (McGuire {\it et al.}\ 2004b).

   The half-human/half-machine `Cyborg' approach (see Figure 1) uses human locomotion for taking the computer vision-algorithms to the field for teaching and testing, using a wearable computer. This is advantageous because we can therefore concentrate on developing the `scientific' aspects for autonomous discovery of features in computer imagery, as opposed to the more `engineering' aspects of using computer vision to guide the locomotion of a robot through treacherous terrain. This means the development of the scientific vision system for the robot is effectively decoupled from the development of the locomotion system for the robot.

   After the maturation of the computer-vision algorithms, we hope to
transplant these algorithms from the Cyborg computer to the on-board
computer of a semi-autonomous robot that will be bound for Mars or one of the interesting moons in our solar system. These algorithms could also work in analyzing remote-sensing date from orbiter spacecraft.

\subsection{Image Segmentation, Uncommon Maps, Interest Maps, and Interest Points}
\markright{The Cyborg Astrobiologist: System: Image Analysis}

With human vision, a geologist, in an unbiased approach to an outcrop (or scene):
\begin{spacing}{1}
\begin{trivlist}
\raggedright
\item$\bullet$ Firstly, tends to pay attention to those areas of a scene which are most unlike the other areas of the scene; and then,
\item$\bullet$ Secondly, attempts to find the relation between the different areas of the scene, in order to understand the geological history of the outcrop.\footnote{This concept can be compared to regular geological base-mapping.}
\end{trivlist}
\end{spacing}

The first step in this prototypical thought process of a geologist was our motivation for inventing the concept of uncommon maps. See  McGuire {\it et al.} (2004b) for an introduction to the concept of an uncommon map, and our implementation of it.
We have not yet attempted to solve the second step in this prototypical thought process of a geologist, but it is evident from the formulation of the second step, that
human geologists do not immediately ignore the common areas of the scene.  Instead, human geologists catalog the common areas and put them in the back of their minds for
``higher-level analysis of the scene'', or in other words, for determining explanations for the relations of the uncommon areas of the scene with the common areas of the scene.

For example, a dark, linear feature transects a light-toned, delineated surface. At this specific scale, the dark feature is uncommon, an ``interest point'', as it has specific relation to the surrounding light-toned material. It can ``tell the story'' of the outcrop. Continued study may show how it cuts the delineation of the light-toned material, most likely indicating a younger age. Coupled with a capacity for microscopic analysis, or even better, spectrographical analysis of mineralogy, a continued study may show the dark feature to be of basaltic composition and the light-toned material to be of granitic composition. This data compared with the information stored in the mind of the geologist (knowledge) may lead to the interpretation of the outcrop as a foliated granite (gniess) cut be a dolerite dike.

 Prior to implementing the `uncommon map', the first step of the prototypical geologist's thought process, we needed a segmentation algorithm, in order to produce pixel-class maps to serve as input to the uncommon map algorithm.
We have implemented the classic co-occurrence histogram algorithm (Haralick, Shanmugan \& Dinstein 1973; Haddon \& Boyce 1990). For this work, we have not included texture information in the segmentation algorithm nor in the uncommon map algorithm. Currently, each of the three bands of Hue, Saturation and Intensity ($HSI$) color information is segmented separately, and later merged in the interest map by summing three independent uncommon maps. In ongoing work, we are working to integrate simultaneous color \& texture image segmentation into the Cyborg Astrobiologist system (e.g., Freixenet, Mu\~noz, Mart\'i \& Llad\'o 2004).

The concept of an `uncommon map' is our invention, though it indubitably has been independently invented by other authors, since it is somewhat useful.\footnote{Note in proofs: News reports in 2005 (i.e., ``Chemical Guidebook May Help Mars Rover Track Extraterrestrial Life'', http://www.sciencedaily.com/releases/2005/05/050504180149.htm ) brought the work at Idaho National Laboratory to our attention, in which the Idaho researchers use a mass spectrometer in raster mode on a sample, in order to make an `image', within which they search for uncommon areas. They also do higher-level fuzzy-logic processing with a Spectral IDentification Inference Engine (SIDIE) of these ``hyperspectral images" of mass spectra. They have capabilities to blast more deeply into their samples, autonomously, if their inference engine suggests that it would be useful.  See Scott, McJunkin, \& Tremblay (2003) and Scott \& Tremblay (2002) for the status of their system as of a couple of years ago.} In our implementation, the uncommon map algorithm takes the top 8 pixel classes determined by the image segmentation algorithm, and ranks each pixel class according to how many pixels there are in each class.  The pixels in the pixel class with the greatest number of pixel members are numerically labelled as `common', and the pixels in the pixel class with the least number of pixel members are numerically labelled as 'uncommon'.  The `uncommonness' hence ranges from 1 for a common pixel to 8 for an uncommon pixel, and we can therefore construct an uncommon map given any image segmentation map.  Rare pixels that belong to a pixel class of 9 or greater are usually noise pixels in our tests thus far, and are currently ignored. In our work, we construct several uncommon maps from the color image mosaic, and then we sum these uncommon maps together, in order to arrive at a final interest map.  

 For more details on our particular software techniques, especially on image segmentation and uncommon mapping, see McGuire {\it et al.} (2004b).

\subsection{Hardware for the Cyborg Astrobiologist}
\markright{The Cyborg Astrobiologist: System: Hardware}

For this mission to Riba, the non-human hardware of the Cyborg Astrobiologist system consisted of:
\begin{spacing}{1}
\begin{trivlist}
\raggedright
\item $\bullet$ a 667 MHz wearable computer (from ViA Computer Systems in Minnesota) with a `power-saving' Transmeta `Crusoe' CPU and 112 MB of physical memory, 
\item $\bullet$ an indoor/outdoor sunlight-readable tablet display with stylus (from ViA Computer Systems),
\item $\bullet$ a SONY `Handycam' color video camera (model {\it DCR-TRV620E-PAL}), and
\item $\bullet$ a tripod for the camera.
\end{trivlist}
\end{spacing}

The SONY Handycam provides real-time imagery to the wearable computer via an IEEE1394/Firewire communication cable. The system as deployed to Riba used two independent batteries: one for the computer, and the other for the camera.  The power-saving aspect of the wearable computer's Crusoe processor is important because it extends battery life, meaning that the human does not need to carry spare batteries. A single lithium-ion battery for the wearable computer, which weighs about 1kg, was sufficient for this 4 hour mission. Likewise, a single lithium-ion battery (SONY model NP-F960, 38.8Wh) was sufficient for the SONY Handycam for the 4 hour mission to Riba, despite frequent use of the power-hungry fold-out LCD display for the Handycam.

The main reason for using the tripod during the mission to Riba is that it allows the user to repoint the camera or to zoom in on a feature in the previously-analyzed image. In the previous study at Rivas (McGuire {\it et al.} 2004b), mosaicking and automated repointing was part of the study, so the stable platform provided by the tripod was essential. We eliminated the Pan-Tilt Unit from this mission to Riba because it adds to the mobility of the Cyborg Astrobiologist system, by eliminating the extra bag containing a battery and communication \& power cables to the Pan-Tilt Unit. Another reason for eliminating the Pan-Tilt Unit from the system for the mission to Riba is that it saves time since there is less waiting around for the mosaicking and repointing to be completed. The capacities of automated mosaicking and of automated repointing at interest points are, however, essential to the system in the long run, and will be re-introduced at a later stage when needed.\footnote{The pan-tilt unit was not absolutely essential for the Riba mission, but it could have been useful. The geologists had some troubles at Riba, when they wanted to use the results of the Cyborg Astrobiologist's analysis of a wide-angle image in order to repoint the camera at an interest point, and then zoom in on it. The geologists at one point decided to have one geologist point with his hand at a desired location on the rather uniformly-colored sandstone outcrop, so that the other geologist who was controlling the pointing of the camera could find that feature in the camera's view-finder and point at the interesting feature and zoom in on it (see Figure~\ref{EnriqueJens}). If we had had the Pan-Tilt Unit with us, then this would have been much more automatic, as the repointing could have been under computer control.} During this mission, the Cyborg Astrobiologist system analyzed 32 images from 7 different tripod positions at 3 different outcrops over a 600 meter distance and over a 4 hour period. During the previous mission at Rivas, with the Pan-Tilt Unit enabled, the Cyborg Astrobiologist system only analyzed 24 mosaic images from 3 different tripod positions at only one outcrop over a 300 meter distance and over a 5 hour period. See Table~\ref{MissionParameters} and Figure~\ref{MissionMap}.

 For this particular study, the mobility granted by the wearable computer was almost essential. Using a more powerful non-wearable computer could have restricted the mobility somewhat, and would have made it more difficult to study the third outcrop on the slopes of the castle-topped hill.  The head-mounted display that we used during the Rivas mission (2004) was much brighter than the tablet display used during the Riba mission. Together with the thumb-operated finger mouse, the head-mounted display was more ergonomic during the mission to Rivas, than was the tablet display and stylus used during this Riba mission. However, the spatial resolution of the head-mounted display was somewhat less than the resolution of the tablet display. During the Riba mission, we wanted to share and interpret the results interactively between the three investigators. This would have been much more difficult with the single-user head-mounted display than it was with the multi-viewer higher-resolution tablet display. So we used the tablet display during the mission to Riba, with the intention of switching to the head-mounted display later in the day.\footnote{The multi-viewer interactivity provided by the tablet display offered more to us than the brighter screen of the head-mounted display, so we used the tablet display for the entire mission to Riba. It would be useful to be able to switch rapidly between the tablet display and the head-mounted display, but our system remains incapable of this at this moment.}

\subsection{Software for the Cyborg Astrobiologist}
\markright{The Cyborg Astrobiologist: System: Software}

 The wearable computer processes the images acquired by the color digital video camera, to compute a map of ``interesting'' areas. What the system determines as ``interesting'' is based on the ``uncommon'' maps (introduced in Section 2.1). It is the relation between uncommon and common that eventually can tell the geological history of the outcrop.  The computations use two-dimensional histogramming for image segmentation (Haralick, Shanmugan \& Dinstein 1973; Haddon \& Boyce 1990).  This image segmentation is independently computed for each of the Hue, Saturation, and Intensity (H,S,I) image planes, resulting in three different image-segmentation maps. These image-segmentation maps were used to compute `uncommon' maps (one for each of the three (H,S,I) image-segmentation maps): 
each of the three resulting uncommon maps gives highest weight to those regions of smallest area for the respective (H,S,I) image planes. Finally, the
three (H,S,I) uncommon maps are added together into an interest map, which is used by the Cyborg system in order to determine the top three interest points to report to the human operator. The image-processing algorithms and robotic systems-control algorithms are all programmed using the graphical programming language, NEO/NST (Ritter {\it et al.} 1992, 2002). Using such a graphical programming language adds flexibility and ease-of-understanding to our Cyborg Astrobiologist project, which is by its nature largely a software project. We discuss some of the details of the software implementation in McGuire {\it et al.} (2004b).

  After segmenting the mosaic image (see Figure~\ref{Explanation}), we use a very simple method to find interesting regions in the image:  we look for those regions in the image that have a significant number of uncommon pixels. We accomplish this by: first, creating an uncommon map based upon a linear reversal of the segment peak ranking; second, adding the 3 uncommon maps (for $H$, $S$, \& $I$) together to form an interest map; and third, blurring this interest map\footnote{with a gaussian smoothing kernel of width $B=10$. This smoothing kernel effectively gives more weight to clusters of uncommon pixels, rather than to isolated, rare pixels.}. 

  Based upon the three largest peaks in the blurred/smoothed interest map, the Cyborg system then shows the human operator the locations of these three positions, overlaid on the original image. The human operator can then decide how to use this information: i.e., whether to ignore the information or to zoom in on one of the interest points. This step can be automated in future versions.

\section{Descriptive Summaries of the Field Site and of the Expedition}
\markright{The Cyborg Astrobiologist: Field Site and Expedition}

   On the 8th of February, 2005, three of the authors (D\'iaz Mart\'inez, Orm\"o \& McGuire) tested the ``Cyborg Astrobiologist" system for the second time at a geological site, with red-sandstone layers, near the village of Riba de Santiuste, north of Sig\"uenza in the northern part of the province of Guadalajara (Spain).

 \begin{figure}[p]
 \center{\includegraphics[width=16cm]{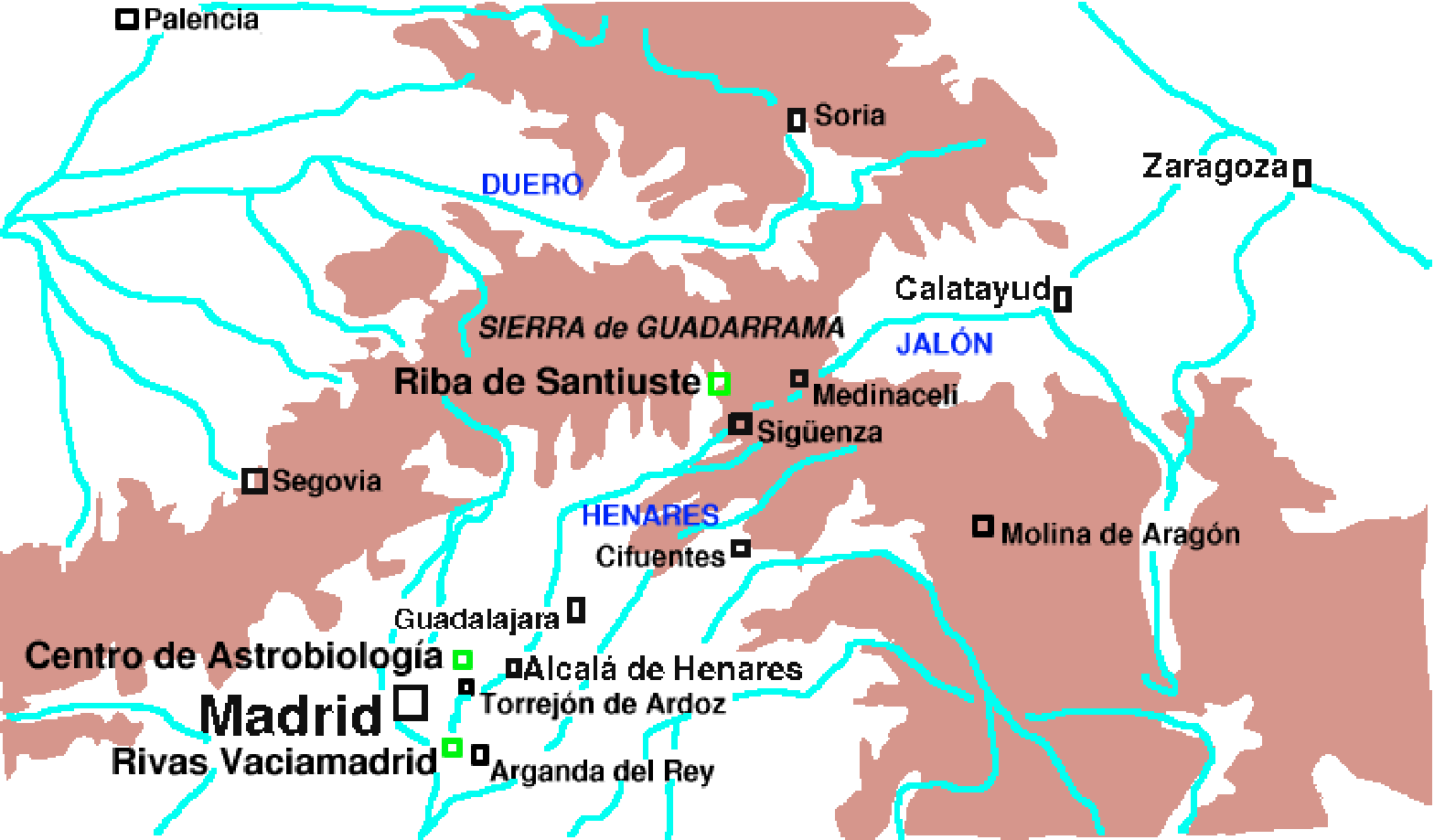}}
  \caption{Map of the north-east part of central Spain, showing the current field site, Riba de Santiuste, and the previous field site, Rivas Vaciamadrid. The location of part of our team's research center, the Centro de Astrobiolog\'ia, is also indicated. The horizontal size of this map is about 520 km.}
 \center{\includegraphics[width=14cm]{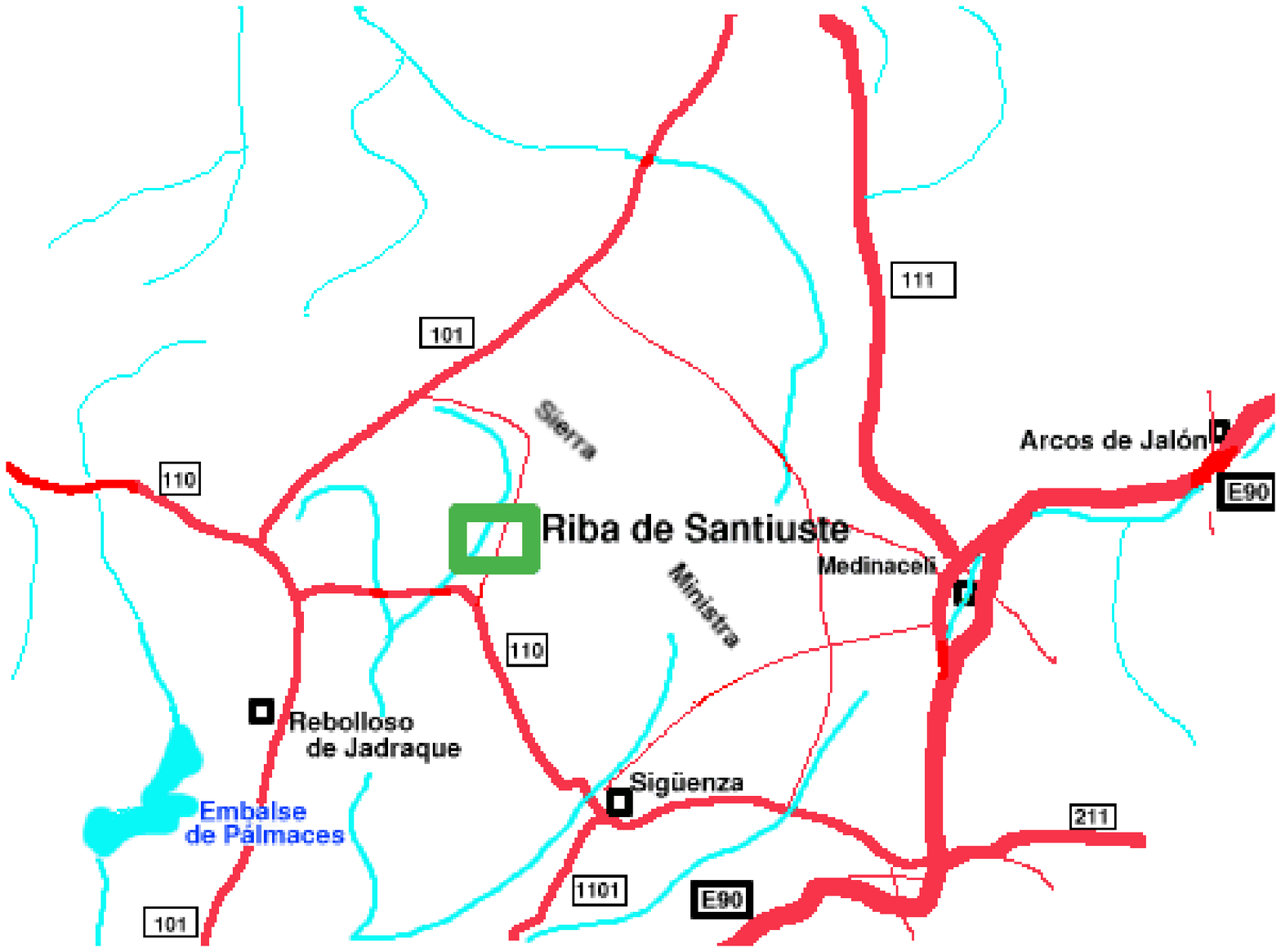}}
  \caption{Map of the vicinity of Riba de Santiuste, with road-numbers indicated. The horizontal size of this map is 100 km.}
 \end{figure}
 
\begin{figure}[th]
 \center{\includegraphics[width=16cm]{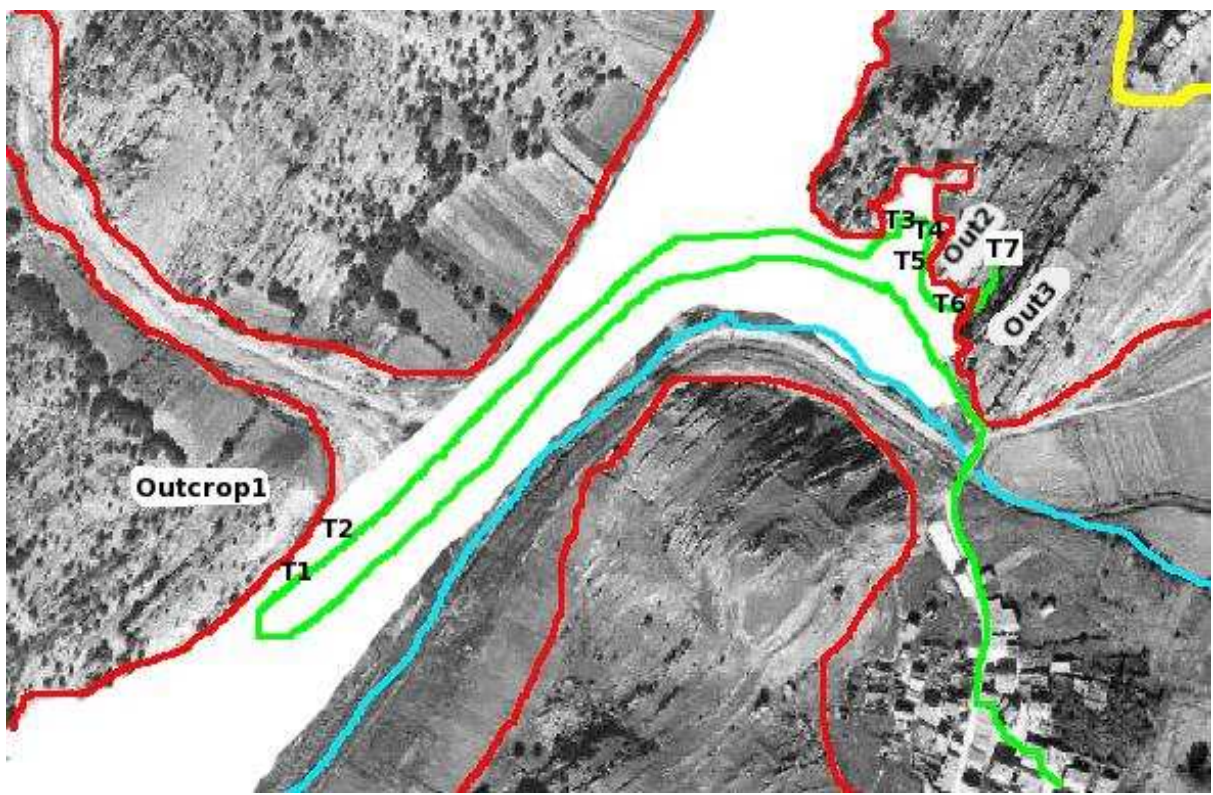}}
  \caption{Map showing the Outcrops (Outcrop1, Out2, Out3) and Tripod Positions (T1-T7) during the February 8, 2005, Cyborg Astrobiologist mission to Riba de Santiuste in Guadalajara.  The village of Riba de Santiuste is shown in the lower right, and the deserted hilltop castle of Riba de Santiuste is indicated by the yellow outline in the upper right. The outcrop ``Outcrop1" had some strata that were colored white, as well as a few small 5-10 cm white spots, and a number of iron-oxide concretions. The outcrop ``Out2" has a whitish-colored mineralization near its base. The outcrop ``Out3'' has alternating layers of red silt-stones and darker-colored paleosols. The approximate paths of the river stream are indicated in cyan, and the approximate path of our mission is indicated by the green line that goes over the small bridge. The central valley of the image has been colored white for clarity. The horizontal size of this aerial photograph is about 1km. }
\label{MissionMap}
 \end{figure}

\begin{table}
\begin{tabular}{| p{2.8cm} | p{4.5cm} | p{4.5cm} | p{3cm} |}
\hline
{\bf Mission Location}    & {\bf Mission Date(s)}              &    {\bf Mission Mode}            & {\bf Geology}            \\ \hline
Rivas Vaciamadrid         &    3 March 2004                    & w Pan-Tilt Unit                  & gypsum cliffs            \\ 
                          &    11 June 2004                    & w Mosaicking                     & gypsum wettening         \\ 
                          &                                    & w Automated Repointing           & Miocene age              \\  
                          &                                    & w/o Zoom Lens                    &                          \\  
                          &                                    & w Head-mounted display           &                          \\  \hline
Riba de Santiuste         &    8 February 2005                 & w/o Pan-Tilt Unit                & ``Red Beds''             \\
                          &                                    & w/o Mosaicking                   & red sandstones           \\ 
                          &                                    & w/o Automated Repointing         & redox geochemistry       \\  
                          &                                    & w Zoom Lens (manual)             & sulfate deposits         \\  
                          &                                    & w tablet display                 & Triassic age             \\  \hline
\end{tabular}
\vspace{1.5cm}

\begin{tabular}{| p{2.8cm} | p{4.5cm} | p{4.5cm} | p{3cm} |}
\hline
{\bf Mission Location}    & {\bf Setup time}                   &  {\bf Mission Duration}          & {\bf Mission Distance}   \\ \hline
Rivas Vaciamadrid         & 1 hr                               &   5 hours                        & 300 m                    \\ 
                          &                                    &                                  & (by foot)                \\  \hline
Riba de Santiuste         & 0.5 hrs                            &   4 hours                        & 600 m                    \\
                          & (w teaching)                       &                                  & (by vehicle, and         \\ 
                          &                                    &                                  & partly by foot)          \\  \hline
\end{tabular}
\vspace{1.5cm}

\begin{tabular}{| p{2.8cm} | p{4.5cm} | p{4.5cm} | p{3cm} |}
\hline
{\bf Mission Location}    &   {\bf Number of Outcrops studied} & {\bf Number of Tripod Positions} & {\bf Images Analyzed}    \\  \hline
Rivas Vaciamadrid         &         1                          &        3                         &     24                   \\  \hline
Riba de Santiuste         &         3                          &        7                         &     32                   \\  \hline
\end{tabular}
\vspace{1.5cm}

\begin{tabular}{| p{2.8cm} | p{4.5cm} | p{3.5cm} | p{3.5cm} |}
\hline
{\bf Mission Location}    &  {\bf \# Interest Points w human-geologist concurrence} & {\bf \# False-Negative Interest Points} & {\bf \# False-Positive Interest Points}    \\  \hline
Rivas Vaciamadrid         &         -- & -- & --                                                \\  \hline
Riba de Santiuste         &         69 & 32 & 32                                            \\  \hline
\end{tabular}
\caption{A description of some of the parameters of the missions to Rivas Vaciamadrid and to Riba de Santiuste.}
\label{MissionParameters}
\end{table}

   These tests at Riba are meant to be a complementary, confirmatory test of the methodology first tested in the spring of 2004 in Rivas Vaciamadrid. The site in Riba may have more direct relevance as an analog for the Terra Meridiani site on Mars, which the Mars Exploration Rover, Opportunity, is now exploring. Riba has red sandstone beds, with some outcrops showing local chemical bleaching due to reduction and mobilization of iron, as well as precipitation of oxidized iron impurities. This process gives the sandstone both dark-red and white colored areas. Within the red sandstone, there are concretions of dark-red oxidized iron, as well as some sites where concretion had only been partially complete. Furthermore, the coloring of the red sandstones at Riba is much more ``brilliant" and saturated than the unsaturated white coloring of the clay- \& sulfate-bearing cliffs in Rivas. The dichotomy and contrast of the colors at Riba, between the oxidized red and the bleached white colors of the sandstones\footnote{Similar imagery and astrobiological hypotheses presented recently by Hofmann (at the European Astrobiology Network Association (EANA) conference in November 2004, at the Open University in Milton Keynes, United Kingdom) inspired us to find such a site in Spain (Hofmann 1990, 2004). Hofmann told one of the authors (McGuire) that there certainly should be several such ``red bed'' sites in Spain. Later, in December 2004, D\'iaz Mart\'inez and McGuire explored several sites in Guadalajara, and found that the site at Riba de Santiuste had some nice red beds, exhibiting the white bleached areas of red sandstone, which we decided to use for this study.}$^,$\footnote{Hofmann refers to the white circularly-shaped zones where the iron has been first reduced and then removed as ``reduction spheroids''. These spheroids only have a reduced chemical state prior to the removal of the iron, which is mobile when it is reduced. He posits a ``mobile and kinetically inert'' biological agent as a possible source of the chemical reduction of the previously-oxidized iron. In his studies, the reduced iron is often removed by concentration at the center of the reduction spheroid. This concentrated iron is later oxidized again, forming a dark-red concretion. Since we only observed a two-dimensional slice (a white disk) through the three-dimensional sphere of the ``reduction spheroid'', this may explain why we did not observe any concretions at the centers of the white circles in the red beds at Riba. The concretions at Riba tended to be in a nearby zone that was physically separate from the white zone in the red beds.}$^,$\footnote{Maybe there have been other concretion-formation mechanisms active at Riba, beyond the model of Hofmann, in which concretions tend to form at the centers of the reduction spheroids. A model, based upon concretion formation in Utah red beds, which includes reducing fluids passing through permeable sandstone beds, has been suggested for the Meridiani hematite deposits on Mars (Orm\"o {\it et al.} 2004, Chan {\it et al.} 2004). In their model, the reducing fluids remove oxidized iron coatings of the sand grains, and later precipitate the iron as spherical concretions when the iron encounters oxidized groundwater.}
certainly made the study of these ``red beds" into a relatively straight-forward, but highly-discriminating, test of the Cyborg Astrobiologist's current computer vision capabilities.

The rocks at the outcrops at Riba are of Triassic age (260-200 Myrs before present), and consist of sandstones, gravel beds, and paleosols. The rocks were originally deposited during the Triassic in different layers by the changing depositional processes of a braided river system. This river system consisted of active channels with fast transport of sand grains and gravel. During different millenia in the Triassic, the river system shifted and evolved. Therefore, in the example of the paleosol outcrop, the deposition was only of fine-grained silt, which later was affected by soil processes and formed the paleosol layers. The rock layers were folded by Alpine tectonics in the Cenozoic. 

   We arrived at the site at 12 noon on February 8, 2005, and in the next 30 mins the roboticist quickly assembled the Cyborg Astrobiologist system, and taught the two geologists how to use the system. For the next 4 hours, the two geologists were in charge of the mission, deciding where to point the Cyborg Astrobiologist's camera, interpreting the results from the tablet display, and deciding how to use the Cyborg Astrobiologist's assessment of the interest points in the imagery. Often the geologists chose one of the top three interest points, and then either zoomed in on that point with the camera's zoom lens, or walked with the camera and tripod to a position closer to the interest point, in order to get a closer look at the interest point with the computer vision system.

 The computer was worn on the geologist's belt, and typically took 30 seconds to acquire and process the image. The images were downsampled in both directions by a factor of 2 during these tests, and the central area was selected by cropping the image; the final image dimensions were $192 \times 144$.    

We chose a set of 7 tripod positions at 3 geological outcrops at Riba de Santiuste in order to test the Cyborg Astrobiologist system (see the map in Figure~\ref{MissionMap}). This is an improvement upon the number of tripod positions and outcrops studied in the first mission to Rivas Vaciamadrid in the spring of 2004 (see Table 1). The first geological outcrop at Riba de Santiuste consisted of layered deposits of red sandstone. In several of the images (see Figures 6A, 6B, \& 6D for examples), the wearable computer determined aspects of white bleaching to be interesting. Furthermore, in several of the images (see Figures 6C \& 6D for examples), the wearable computer found the concretions to be interesting. The second geological outcrop was partly covered by a crust of sulfate and carbonate minerals. At this outcrop, the computer vision system found the rough texture of the white-colored mineral deposits to be interesting (see Figure 7A). At the third geological outcrop, we pointed the camera and computer vision system at a paleosol. The computer vision system of the wearable computer found the calcified root structures within the paleosols to be of most interest (see Figure 7B).

\section{Results}  
\markright{The Cyborg Astrobiologist: Results}
Without the problems with shadow and texture that we encountered during our first mission at the white gypsum-bearing cliffs of Rivas Vaciamadrid, the system performed admirably. See Figure 6B for such an example of an image without shadow or texture. We show more details of the automated image processing by the wearable computer of this image in Figure 5. This image contained an area with bleached domains (in white) in the red sandstone, as well as a few isolated iron-oxide concretions of  dark red color. Processing this image by image segmentation and by uncommon-map construction was straightforward. The wearable computer reported to the geologists that the three most interesting points were two areas of the white bleached domain, and a third point in the lower part of the image which was a somewhat-darker tone of red. The geologists also found the smaller dark-red concretions in the upper part of the image to be interesting. The wearable computer found these smaller dark-red concretions to be interesting, but only before the interest map was smoothed (or blurred), as summarized in Figure 5. The wearable computer does not report the top three points from the unblurred interest map to the user. However, the full unblurred interest map, as well as the blurred interest map, are both displayable to the user in real time, and they are both stored to disk for post-mission analysis. We regard the fact that the wearable computer had interest in the dark red concretions, even though it did not report this interest in its top 3 interest points, as a partial success. Perhaps, if the camera had zoomed in by a factor of 10, it would have found these dark red concretions to be interesting, even in the blurred interest map. Hence, these features would be noticed when further approaching the outcrop. Even at this distance, the small size of the dark red concretions was at the limits of the perception capabilities of the geologists themselves.  Nonetheless, the computer did inform the geologists that it had interest in the white bleached spots and the darker red region at the bottom of the image, which we regard to be a higher level of success.

In Figures 6 \& 7, we present a number of other images that were acquired and processed by the Cyborg Astrobiologist at Riba de Santiuste.
For each image, we show here the image segmentation for the saturation slice of the color image, since the saturation slice discriminated rather
well the red and white colors. We also show the final, blurred interest map, which was computed by summing the uncommon maps for Hue, Saturation \& Intensity as in Figure 5, and then smoothing the resulting map.
Figure 6A shows the border between a large oxidized red-colored zone of the red bed and a large bleached white-colored zone of the red bed. 
The wearable computer found the texture in the white zone to be most interesting.
Figure 6C shows a highly-magnified view of some of the concretions in the red bed near Tripod position 2. These concretions are not well-rounded like the ``blueberries''; rather, they are more like the ``popcorn'' concretions observed by Opportunity at Meridiani Planum. The computer vision system in the wearable computer found two of the darker-red concretions on the left side of the image to be interesting, and one bright white area on the right side of Figure 6C to be interesting. The discriminatory power of the image segmentation was not very clean at this high level of magnification. Figure 6D shows a zoomed-out view of some concretions and a bleached white zone near Tripod position 2. The computer vision system found the area of the concretions to be most interesting, followed by the bleached area in the left side of the image. As shown in Figure 6E, in the same part of the outcrop were some quartzite rocks of green, gray and yellow colors. In an image containing these rocks, the computer vision system found several of them to be more interesting than the surrounding red sandstone.

Likewise, at the second outcrop, we observed some highly-textured mineral deposits of whitish colors.
In a zoomed-out view of these mineral deposits (Figure 7A), the wearable computer found the texture of the mineral deposits to be more interesting than the more-homogenous red color of the neighboring sandstone.\footnote{The software of the Cyborg Astrobiologist has not been configured to segment the images into different regions, based upon the texture of each different region. The software only uses gray-level differences for the image segmentation, but for each of the 3 different color layers. However, when there is a significant amount of variation of the gray-levels within a region (which is often caused by texture), then the software will likely label many of the pixels in that region as being uncommon. Therefore, our system does have some capabilities to study texture.}
The wearable computer did not find the highly-textured region containing many dark holes to be as interesting as the white-textured mineral deposits.
In an ideal computer vision system, one of the top three interest points probably should have been within the region with all of the dark holes. These dark holes formed by a combination of physical (wind erosion) and chemical (differential cementation) processes.\footnote{The current Cyborg Astrobiologist system certainly is biased towards color or intensity differences over differences in physical morphology. Our system uses color properties, and not morphology to make its decisions. In geological samples, the color differences are often a result of differences in chemistry (i.e., differences in chemical composition, state of oxidation of iron, {\it etc.}). Furthermore, in geological samples, the differences in physical morphology also depend on chemistry (erodability, which in turn depends on differential cementation, {\it etc.}), but mostly on the geometry of internal structures, their size, spacing, {\it etc.} Certainly, adding morphology as a layer in the interest map would be very beneficial to our system.}  Based partly upon the guidance of the wearable computer, we acquired several samples of the mineral deposits for analysis in the laboratory. Based upon the geologists' experience and upon acid tests of these mineral deposits at the field site, we suspected that the mineral deposits consisted largely of a white textured overlayer of 15-20 mm of a sulfate, with a thin gray underlayer of 1-10 mm of calcite. The laboratory tests later confirmed that the overlayer is composed of largely of hydrated calcium sulfate (gypsum).

At the third outcrop, the Cyborg Astrobiologist system studied a layer of paleosols, which are ``affected'' silty deposits from a fluvial plain.\footnote{``Paleosol'' is the series of features (color, structures, layers, minerals, {\it etc.}) which indicate that the sediment was affected by {\it soil processes} in the past. Such soil processes include leaching, oxidation, reduction, and precipitation.}
These paleosols contained partly-calcified root structures of plants that grew in this fluvial plain during the Triassic. At Tripod position 6,
 our computer vision system found the white-colored calcified-root structures to be more interesting than the surrounding matrix of red siltstone (Figure 7B). However, at Tripod position 7, in a more complex image (not shown here) with more root structures and more diverse coloring, the computer vision system did not find the root structures to be particularly interesting.\footnote{At Tripod position 6, the system worked much better because the root structures were a sufficiently minor component of the image; the reddish-colored parts of the paleosol dominated. At Tripod position 7, the system did not work well because the root structures covered an area in the image which was greater than the areas of some of the other minor components of the image. In future enhancements of the computer vision system, if root structures are deemed by the mission scientists to be of sufficient interest, then we could develop an interest-map algorithm that would respond to webs of linear connected features like root structures, or we could develop a different interest-map algorithm which could be biased towards whitish colored areas which are unlike all the other colors of the mostly-reddish paleosols.}

 We have compared the Cyborg Astrobiologist's performance in picking the top 3 interest points with a human geologist's quasi-blind classification. Geologist D\'iaz Mart\'inez verbally noted the interesting parts of the image. Then D\'iaz Mart\'inez was shown the Cyborg Astrobiologist's top three interest points, and he then judged how well the computer vision system matched his own image analysis. We judged that there was concurrence between the human geologist and the Cyborg Astrobiologist on the interest points about 69 times (true positives), with 32 false positives and with 32 false negatives, for a total of 32 images studied. This was an average of 2.1 true positives (TP) for each image out of a possible (usually) 3 trials per image; an average of 1.0 false positives (FP) per image out of a possible (usually) 3 trials per image; and an average of 1.0 false negatives (FN) per image, with no {\it a priori} limitation on the number of false negatives per image. There was double counting of positives since sometimes a localized physical feature corresponded to more than one interesting features in the image (i.e. one image had some parallel lines at a geological contact, so the interest points corresponding to this region counted doubly). One way to look at this data is to take the ratios $tpr = TP/(FP+TP)$, $fpr = FP/(FP+TP)$ and $fnr = FN/(FP+TP)$. With our technique, it is difficult to assess the true negative rate $tnr$. For the data from Riba, we compute: a true positive rate of $tpr = 68\%$, a false positive rate of $fpr = 32\%$, and a false negative rate of $fnr = 32\%$. We did not attempt to compute the ROC curves for these images, as a function of the number of interest points computed for each image. A more careful analysis could be made with many more images from different field sites, but these numbers agree qualitatively with the results we were getting at Rivas Vaciamadrid, even though we have not computed the statistics.  Qualitatively, we expect that for the types of imagery that we typically study, the Cyborg Astrobiologist will not do a near-perfect job (with geologist-concurrence analysis giving  $fpr$ and $fnr$ both less than 10\%), nor will it do a poor job (with geologist-concurrence analysis giving either or both $fpr$ and $fnr$ greater than 90\%), it typically has mid-level false-positive and false-negative rates -- a mid-level in performance. Of course, we would like to improve the system to do a near-perfect job, but considering the simplicity of the image-segmentation algorithm and the uncommon-mapping technique, we believe that this work can provide a good basis for further studies.

 One approach for improving the false positive rate is to include further filtering of the interest points that are currently being determined from the uncommon maps. That way, errant positives could be caught before they are reported to the human operator. Right now in the field, the human operator serves as a decent filter of false positives, so fixing the false positive rate is not so urgent; fixing the false negative rate is somewhat more important. One approach for improving the false negative rate would be to compute interest features in new and different ways (i.e., with edge detectors or parallel-line detectors), in order to ensure that all the interesting features are detected.  Both approaches could be attacked, for example, by using context-dependent geologist knowledge, which would need to be coded into a future, more advanced Cyborg Astrobiologist.  Currently, considering that we do not deploy such context-dependent geologist knowledge, our relatively-unbiased uncommon-mapping technique seems to be doing rather well.

\begin{figure}[th]
\center{\includegraphics[width=16cm]{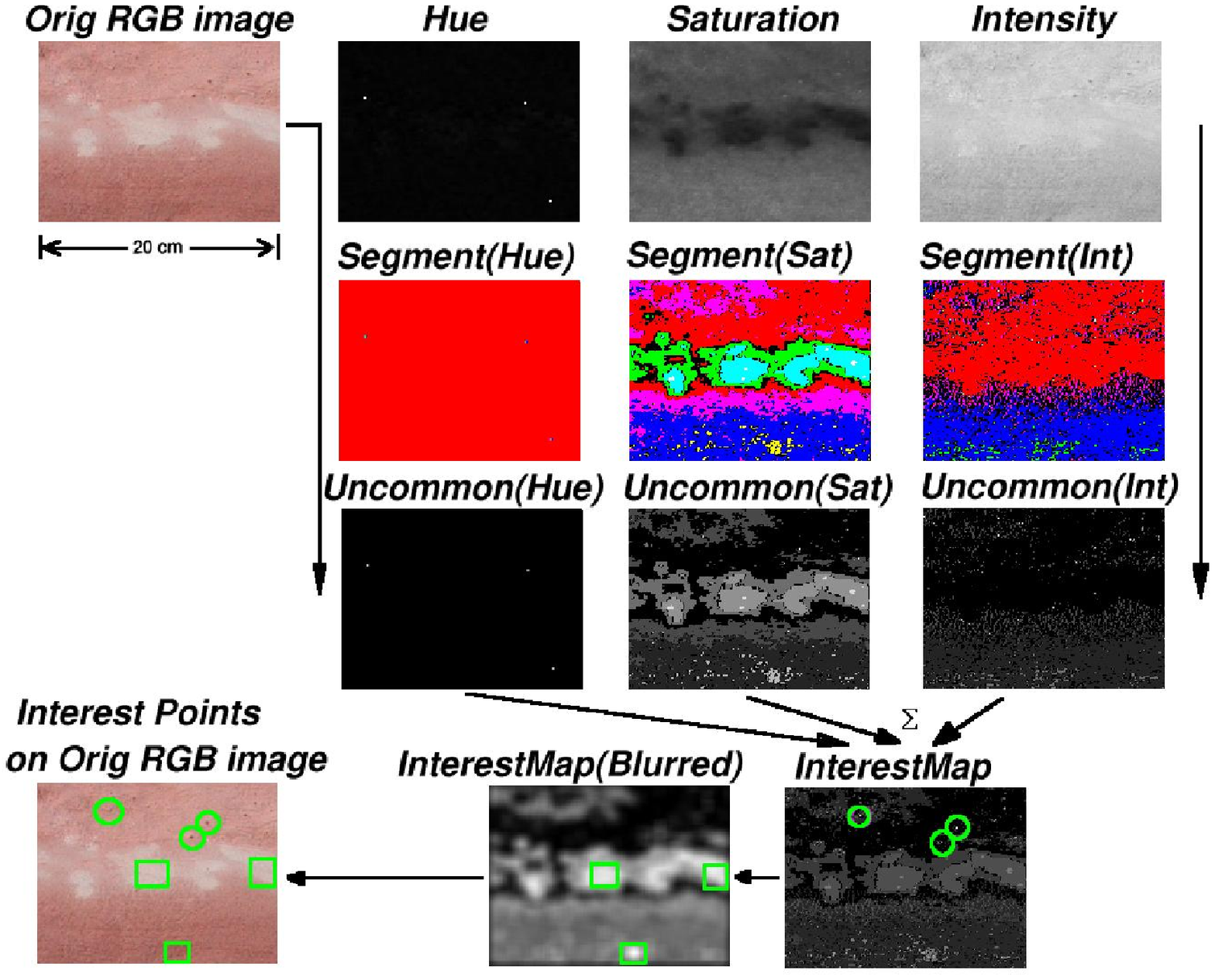}}
\caption{An image of some white spots and nodules, acquired and processed at tripod position \#2 at Riba de Santiuste on February 8, 2005. The wearable computer's autonomous processing of the image is shown, with separate image segmentation and uncommon mapping for each of the hue, saturation and intensity components. The system indicated to the user that the three points indicated by green squares were the most interesting points on a 10 pixel scale. The system also computed that the three points indicated by the green circles were the most interesting points without any spatial summing (on a 1 pixel scale), but it did not indicate this to the user in the field.}
\label{Explanation}
\end{figure}

 \begin{figure}[th]
\center{\includegraphics[width=16cm]{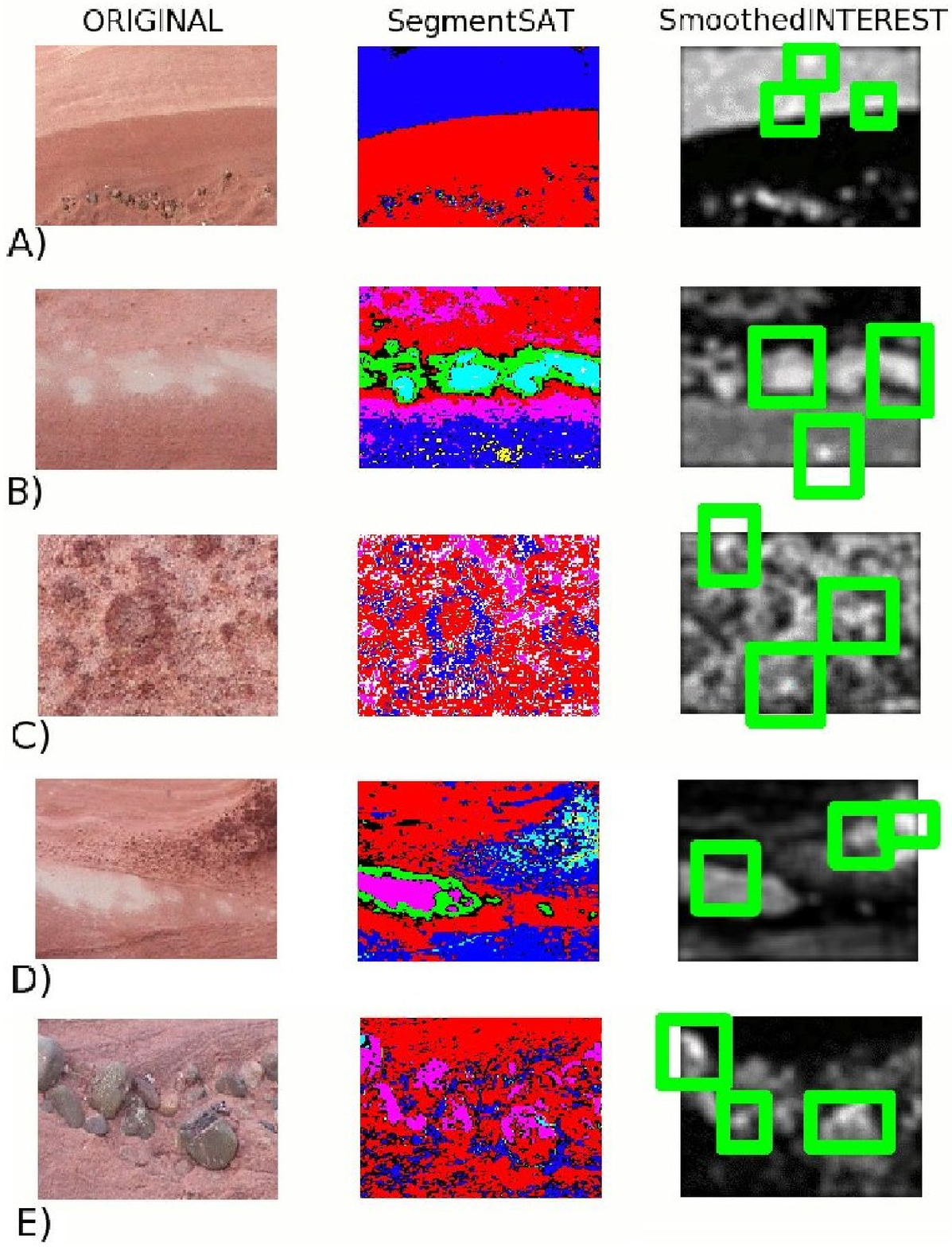}}
  \caption{Figure showing some examples of various images, their image segmentations (for saturation), their interest maps, and their top 3 interest points. These images were acquired and processed at Tripod positions 1 \& 2, near Outcrop \#1.}
 \end{figure}

 \begin{figure}[th]
\center{\includegraphics[width=16cm]{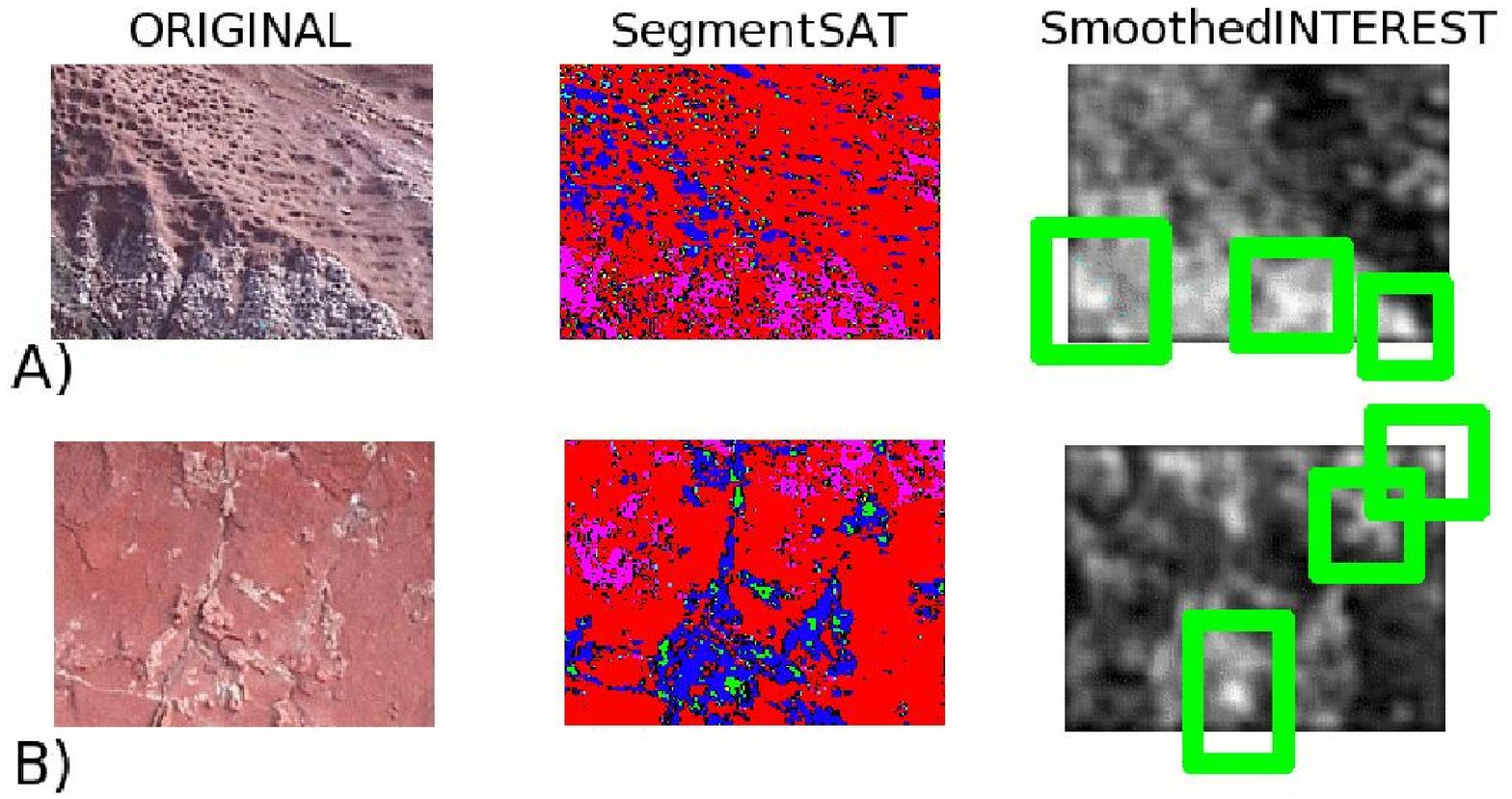}}
  \caption{Figure showing some examples of various images, their image segmentations (for saturation), their interest maps, and their top 3 interest points. These images were acquired and processed at Tripod positions 4 \& 6, near Outcrops \#2 \& \#3.}
 \end{figure}

\section{Discussion \& Conclusions}

We have shown that our Cyborg Astrobiologist exploration system performs reasonably well at a second geological field site. Given similar performance at the first geological field site, we can have some degree of confidence in the general unbiased approach towards autonomous exploration that we are now using. We can also have sufficient confidence in our general technique that we can use our specific technique of image segmentation and uncommon mapping as a basis for further algorithm development and testing.

In the near future, we plan to:
\begin{itemize}
\item Upgrade our image-segmentation algorithm, in order to handle texture and color simultaneously (cf. Freixenet {\it et al.} 2004). This upgrade may give us more capabilities to handle shadow as well.
\item Test the Cyborg Astrobiologist system at field sites with a microscopic imager. This will complete our mimicry of a stepwise approach by a human geologist towards an outcrop.
\end{itemize}

At Riba de Santiuste, our system has been tested for exploration on imagery at a site that is not unlike Meridiani Planum, where the MER Opportunity is now exploring. Riba de Santiuste is similar to Meridiani Planum in its iron-oxide concretions and its sulfate mineral deposits. The bleached-white zones of the red sandstones at Riba de Santiuste may be an analog for similar geochemical or perhaps biological phenomena on Mars (Chan {\it et al.} 2004; Orm\"o {\it et al.} 2004; Hofmann 1990, 2004). The software behind the Cyborg Astrobiologist's computer vision system has shown its initial capabilities for imagery of this type, as well as for imagery of some other types. This computer vision software may one day be mature enough for transplanting into the computer-vision exploration system of a Mars-bound orbiter, robot or astronaut.

%\begin{acknowledgements}
\section{Acknowledgements}

P. McGuire, J. Orm\"o and E. D\'iaz Mart\'inez would all like to thank the Ramon y Cajal Fellowship program of the Spanish Ministry of Education and Science. Many colleagues have made this project possible through their technical assistance, administrative assistance, or scientific conversations. 
We give special thanks to Kai Neuffer, Antonino Giaquinta, Fernando Camps Mart\'inez, and Alain Lepinette Malvitte for their technical support.
We are indebted to Gloria Gallego, Carmen Gonz\'alez, Ramon Fern\'andez, Coronel Angel Santamaria, and Juan P\'erez Mercader for their administrative support.
We acknowledge conversations with Beda Hofmann, Virginia Souza-Egipsy, Mar\'ia Paz Zorzano Mier, Carmen C\'ordoba Jabonero, Josefina Torres Redondo, V\'ictor R. Ruiz, Irene Schneider, Carol Stoker, J\"org Walter, Claudia Noelker, Gunther Heidemann, Robert Rae, Jonathan Lunine, Ralph Lorenz, Goro Komatsu, Nick Woolf, Steve Mojzsis, David P. Miller, Bradley Joliff, Raymond Arvidson, and Daphne Stoner.  The field work by J. Orm\"o was partially supported by grants from the Spanish Ministry of Education and Science (AYA2003-01203 and CGL2004-03215). The equipment used in this work was purchased by grants to our Center for Astrobiology from its sponsoring research organizations, CSIC and INTA.

%\end{acknowledgements}

\makeatletter	
\renewcommand{\@biblabel}[1]{}
\makeatother

\end{spacing}

\markright{The Cyborg Astrobiologist: Scouting Red Beds in Guadalajara}


\begin{thebibliography}{}
\markright{The Cyborg Astrobiologist: References}


\bibitem{}
Chan, M.A., Beitler, B., Parry, W.T., Orm\"o, J. \& Komatsu, G. (2004). 
A possible terrestrial analogue for haematite concretions on Mars.
{\it Nature} {\bf 429}, 731-734. 

\bibitem{}
Compton, W.D. (1989). {\it Where no man has gone before: A History of Apollo Lunar exploration missions}, chapters 13-14.
Based on NASA History Series {\it SP-4214}.\\ http://www.apolloexplorer.co.uk/default.asp?libsrc=/books/sp-4214/cover.htm

\bibitem{}
Freixenet, J., Mu\~noz, X., Mart\'i, J. \& Llad\'o, X. (2004). Color texture segmentation by region-boundary cooperation. {\it Computer Vision -- ECCV 2004, Eighth European Conference on Computer Vision, Proceedings, Part II, Lecture Notes in Computer Science} Springer, Prague, Czech Republic, Eds. Tom\'as Pajdla, Jir\'i Matas, {\bf 3022}, pp. 250-261. 
Also available in the {\it CVonline} archive:\\ http://homepages.inf.ed.ac.uk/rbf/CVonline/LOCAL\_COPIES/FREIXENET1/eccv04.html

\bibitem{}
Haddon, J.F. \& J.F. Boyce. (1990).
Image segmentation by unifying region and boundary information.
{\it IEEE Trans. Pattern Anal.\ Mach.\ Intell.}\ {\bf 12} (10), pp. 929-948.

\bibitem{}
Haralick, R.M.,  Shanmugan, K. \&  Dinstein, I. (1973). Texture features for image classification. {\it IEEE SMC-3} (6), pp. 610-621.

\bibitem{}
Hofmann, B. (1990). Reduction spheroids from northern Switzerland: Mineralogy, geochemistry and genetic models. {\it Chemical Geology} {\bf 81}, pp. 55-81.

\bibitem{}
Hofmann, B. (2004). Redox boundaries on Mars as sites of microbial activity. {\it IV European Workshop on Exo/Astrobiology}, held at the Open University, Milton Keynes, United Kingdom, abstract, \\
http://physics.open.ac.uk/eana/TALKS/Redox\%20boundaries\%20on\%20Mars\%20as\%20sites\%20of\%20microbial\%20activity.pdf

\bibitem{}
McGuire, P.~C., Rodr\'iguez-Manfredi, J.~A., Sebasti\'an-Martinez, E., Gomez-Elvira, J., D\'iaz-Mart\'inez, E., Orm\"o, J., Neuffer, K., Giaquinta, A., Camps-Mart\'inez, F., Lepinette-Malvitte, A., P\'erez-Mercader, J., Ritter, H., Oesker, M., Ontrup, J. \& Walter, J.\ (2004a). Cyborg systems as platforms for computer-vision algorithm-development for astrobiology, {\it Proceedings of the III European Workshop on Exo/Astrobiology}, held at the Centro de Astrobiologia, Madrid, {\it European Space Agency Special Publication, ESA SP-545},  pp. 141-144. http://arxiv.org/abs/cs.CV/0401004

\bibitem{}
McGuire, P.~C., Orm\"o, J.~O., Diaz-Martinez, E., Rodriguez-Manfredi, J.~A., Gomez-Elvira, J., Ritter, H., Oesker, M., \& Ontrup, J.\ (2004b). The Cyborg Astrobiologist: First field experience, {\it International Journal of Astrobiology}, vol. 3, issue 3, pp. 189-207. http://arxiv.org/abs/cs.CV/0410071 

\bibitem{}
Orm\"o, J., Komatsu, G., Chan, M.~A., Beitler, B., and Parry, W.~T. (2004). Geological features indicative of processes related to the hematite formation in Meridiani Planum and Aram Chaos, Mars: a comparison with diagenetic hematite deposits in southern Utah, USA, {\it Icarus} {\bf 171}, pp. 295-316.

\bibitem{}
Ritter, H.,  {\it et al.}\ (1992, 2002).
The Graphical Simulation Tookit, Neo/NST. 
http://www.TechFak.Uni-Bielefeld.DE/techfak/ags/ni/.

\bibitem{}
Schmitt, H.H. (1987). A field geologist's return to the moon, {\it Abstracts of the Lunar and Planetary Science Conference XVIII} {\bf 18}, pp. 880-881.

\bibitem{}
Scott, J.~R., McJunkin, T.~R., Tremblay, P.~L. (2003). Automated Analysis of Mass Spectral Data Using Fuzzy Logic Classification, {\it Journal of the Association for Laboratory Automation} {\bf 8} (2), pp. 61-63.

\bibitem{}
Scott, J.R. \& Tremblay, P.L. (2002).  Highly reproducible laser beam scanning device for an internal source laser desorption microprobe Fourier transform mass spectrometer, {\it Review of Scientific Instruments} {\bf 73} (3), pp. 1108-1116. 

\bibitem{}
Squyres, S., Grotzinger, J.~P.,  {\it et al.}\ (2004). In situ evidence for an ancient aqueous environment at Meridiani Planum, Mars,
{\it Science} {\bf 306}, pp. 1709-1723.

\end{thebibliography}
\end{document}